\def\paperID{0000}
\def\confName{CVPR}
\def\confYear{2025}

\def\paperTitle{Generalized Trajectory Scoring for End-to-end Multimodal Planning}

\def\authorBlock{
    Zhenxin Li\textsuperscript{1, 2} \quad
    Wenhao Yao\textsuperscript{2} \quad
    Zi Wang\textsuperscript{1} \quad 
    Xinglong Sun\textsuperscript{1}\quad
    Joshua Chen\textsuperscript{1} \\
    Nadine Chang\textsuperscript{1} \quad
    Maying Shen\textsuperscript{1} \quad
    Zuxuan Wu\textsuperscript{2} \quad
    Shiyi Lan\textsuperscript{1} \quad
    Jose M. Alvarez\textsuperscript{1} \\
  {$^1$NVIDIA} \quad {$^2$Fudan University}
}

\newif\ifreview 
\newif\ifarxiv \newcommand{\arxiv}{\arxivtrue}
\newif\ifcamera 
\newif\ifrebuttal 

\arxiv 
\pdfoutput=1
\documentclass[10pt,twocolumn,letterpaper]{article}
\usepackage{array, makecell}
\usepackage{tabularx}
\usepackage{booktabs}

\ifreview \usepackage[review]{cvpr} \fi
\ifarxiv \usepackage[pagenumbers]{cvpr} \fi
\ifrebuttal \usepackage[rebuttal]{cvpr} \fi
\ifcamera \usepackage{cvpr} \fi

\usepackage{graphicx}	
\usepackage{amsmath}	
\usepackage{amssymb}	
\usepackage{booktabs}
\usepackage{times}
\usepackage{microtype}
\usepackage{epsfig}
\usepackage[table,xcdraw,dvipsnames]{xcolor}
\usepackage{caption}
\usepackage{float}
\usepackage{placeins}
\usepackage{color, colortbl}
\usepackage{stfloats}
\usepackage{enumitem}
\usepackage{tabularx}
\usepackage{xstring}
\usepackage{multirow}
\usepackage{xspace}
\usepackage{url}
\usepackage{subcaption}
\usepackage{xcolor}
\usepackage[hang,flushmargin]{footmisc}

\ifcamera \usepackage[accsupp]{axessibility} \fi

% ...

\ifarxiv  \fi

\newcommand{\hydra}{GTRS}

\newcommand{\R}[1]{{%
    \textbf{%
        \ifstrequal{#1}{1}{\textcolor{red}{R#1}}{%
        \ifstrequal{#1}{2}{\textcolor{blue}{R#1}}{%
        \ifstrequal{#1}{3}{\textcolor{magenta}{R#1}}{%
        \ifstrequal{#1}{4}{\textcolor{teal}{R#1}}{%
                           \textcolor{cyan}{R#1}%
        }}}}%
    }%
}}

\usepackage{xr-hyper}

\makeatletter
\newcommand*{\addFileDependency}[1]{
  \typeout{(#1)}
  \@addtofilelist{#1}
  \IfFileExists{#1}{}{\typeout{No file #1.}}
}

\makeatother
\newcommand*{\myexternaldocument}[1]{
    \externaldocument{#1}
    \addFileDependency{#1.tex}
    \addFileDependency{#1.aux}
}

\definecolor{cvprblue}{rgb}{0.21,0.49,0.74}
\usepackage[pagebackref,breaklinks,colorlinks,citecolor=cvprblue]{hyperref}
\usepackage[capitalize]{cleveref}
\crefname{section}{Sec.}{Secs.}
\crefname{table}{Table}{Tables}
\crefname{figure}{Fig.}{Figs.}

\ifarxiv \crefname{appendix}{App.}{Apps.}
\else \crefname{appendix}{Suppl.}{Suppls.} \fi

\unless\ifarxiv \myexternaldocument{_supplementary} \fi

\begin{document}
\title{\paperTitle}
\author{\authorBlock}
\maketitle

\begin{abstract}
End-to-end multi-modal planning is a promising paradigm in autonomous driving, enabling decision-making with diverse trajectory candidates. 
A key component is a robust trajectory scorer capable of selecting the optimal trajectory from these candidates.
While recent trajectory scorers focus on scoring either large sets of static trajectories or small sets of dynamically generated ones, both approaches face significant limitations in generalization. 
Static vocabularies provide effective coarse discretization but struggle to make fine-grained adaptation, while dynamic proposals offer detailed precision but fail to capture broader trajectory distributions.
To overcome these challenges, we propose \hydra{} (Generalized Trajectory Scoring), a unified framework for end-to-end multi-modal planning that combines coarse and fine-grained trajectory evaluation. \hydra{} consists of three complementary innovations: (1) a diffusion-based trajectory generator that produces diverse fine-grained proposals; 
(2) a vocabulary generalization technique that trains a scorer on super-dense trajectory sets with dropout regularization, enabling its robust inference on smaller subsets; 
and (3) a sensor augmentation strategy that enhances out-of-domain generalization while incorporating refinement training for critical trajectory discrimination.
As the winning solution of the Navsim v2 Challenge, \hydra{} demonstrates superior performance even with sub-optimal sensor inputs, approaching privileged methods that rely on ground-truth perception.
Code will be available at \url{https://github.com/NVlabs/GTRS}.

\end{abstract}

\vspace{-0.2in}
\section{Introduction}
\label{sec:intro}

End-to-end multi-modal planning has emerged as a powerful approach in autonomous driving.
Unlike traditional uni-modal planners that predict a single trajectory~\cite{chitta2022transfuser, hu2023planning, jiang2023vad}, multi-modal approaches~\cite{dauner2023parting, biswas2024quad, li2025finetuning, chen2024vadv2, li2024hydra, liao2024diffusiondrive, li2025hydrapp} generate multiple candidates, enabling greater adaptability during inference.
This adaptability supports a wide range of applications, including responding to language instructions~\cite{sima2024drivelm, wang2024omnidrive, renz2024carllava, renz2025simlingo}, 
accommodating different driving styles~\cite{li2025finetuning, tang2025hip}, 
and navigating complex driving environments~\cite{li2024hydra, li2025hydrapp, sima2025centaur}.

The typical problem of end-to-end multi-modal planning involves evaluating multiple trajectory proposals through scoring~\cite{ li2024hydra, biswas2024quad, chen2024vadv2, liao2024diffusiondrive, li2025hydrapp, sima2025centaur} given raw sensor data, without access to ground-truth perception.
The planner selects the trajectory with the highest likelihood as the decision.

Current trajectory scoring methods generally fall into two categories:
(1) scoring a large static vocabulary~\cite{chen2024vadv2, li2024hydra, li2025hydrapp, sima2025centaur}, and (2) scoring a small set of dynamically generated proposals~\cite{liao2024diffusiondrive, xing2025goalflow}.
Both approaches face challenges in generalization.
Fixed trajectory vocabularies~\cite{chen2024vadv2, li2024hydra, li2025hydrapp} offer limited flexibility, as they cannot adapt to situations where dynamic proposals are needed.
Meanwhile, methods that rely on a small number of dynamic proposals~\cite{liao2024diffusiondrive, xing2025goalflow} often fail to generalize to unseen trajectories, since the scorer is only exposed to a narrow subset during training.
Ideally, a robust scorer should generalize across diverse trajectory distributions—whether static or dynamic—to handle the full complexity of real-world scenarios.

\begin{figure*}[tp]
    \centering
    \includegraphics[width=\linewidth]{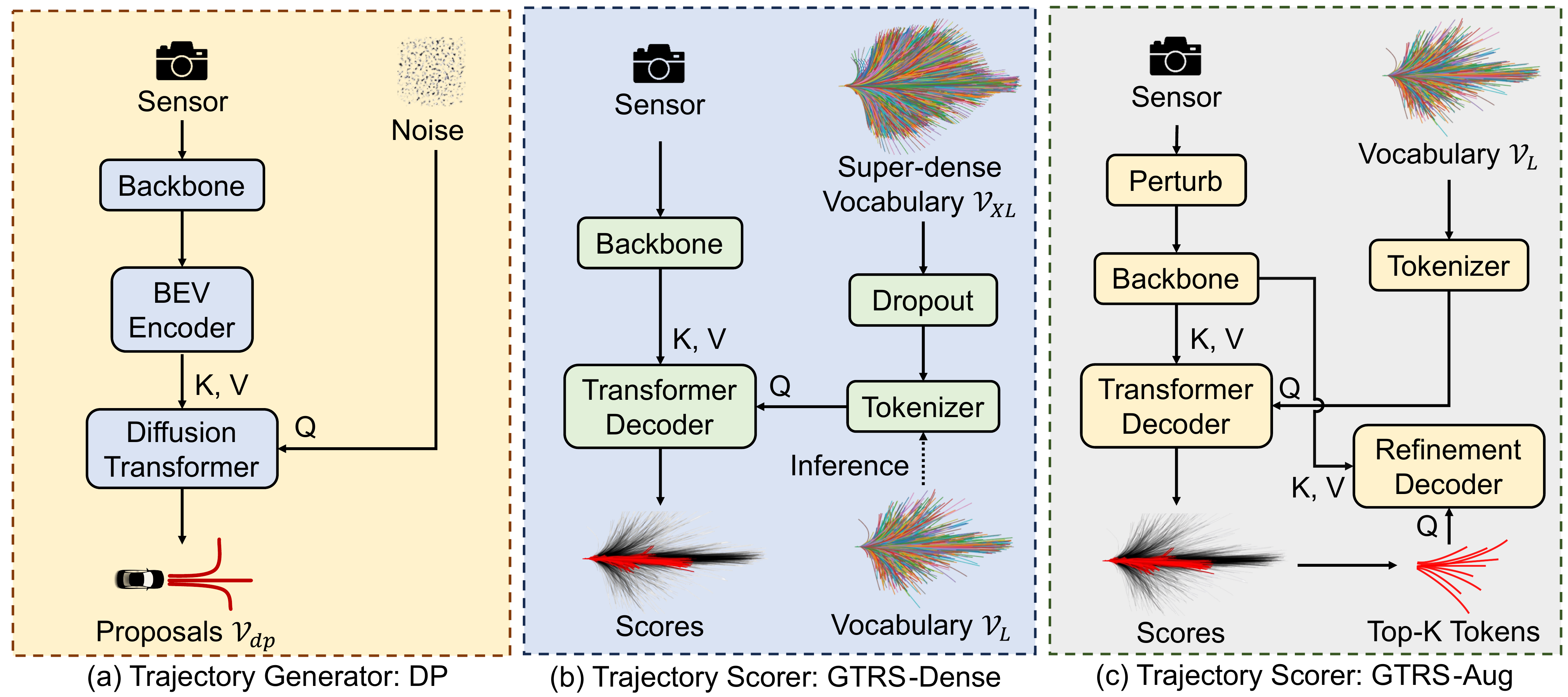}
    \vspace{-0.25in}

    \caption{\textbf{The Three Pillars of \hydra.}}
    \vspace{-0.25in}

    \label{fig:arch}
\end{figure*}

To address these limitations, we propose \hydra{} (Generalized Trajectory Scoring) for end-to-end multi-modal planning.
\hydra{} is built on a key insight: an effective trajectory scorer must be trained on both coarse and fine-grained trajectory distributions to develop robust generalization capabilities. 
Our approach contains three complementary techniques, each leading to a dedicated sub-network as shown in Fig.~\ref{fig:arch}:
\begin{enumerate}
    \item \textbf{Diffusion-based Trajectory Generation (DP)}: 
    A diffusion policy (DP)~\cite{chi2023diffusion} produces diverse trajectory candidates with BEV features as the condition.
    DP provides the fine-grained details crucial for safety-critical situations that coarse, fixed vocabularies cannot capture. (Sec.~\ref{subsec:dp})
    
    \item \textbf{Trajectory Vocabulary Generalization (\hydra-Dense)}: We train on a super-dense vocabulary of trajectory samples (16,384 trajectories) covering a wide range of driving scenarios.
    To maximize generalization of fixed trajectory vocabularies, we propose a trajectory dropout training strategy. 
    The idea is to deliberately create a mismatch between training and inference vocabularies—training the model to effectively generalize to unseen trajectory distributions during inference. (Sec.~\ref{subsec:dense})
    
    \item \textbf{Sensor Augmentation with Refinement (\hydra-Aug)}: 
    To handle unexpected trajectory events and data distribution shifts in the form of viewpoint changes,
    we introduce a data augmentation strategy by applying rotation perturbations to sensor inputs, dramatically improving robustness to out-of-domain environments. 
    Further, a refinement training mechanism enables the model to distinguish between subtly different trajectory options. (Sec.~\ref{subsec:aug})
\end{enumerate}

\hydra{} demonstrates strong trajectory scoring abilities even under sub-optimal sensor conditions, as evaluated on the Navhard benchmark. 
Our contributions are as follows:
\begin{enumerate}
    \item We propose \hydra{}, a generalizable end-to-end multi-modal planning framework that combines diffusion-based trajectory generation with vocabulary scoring. 
    With super-dense vocabularies, sensor augmentations, and refinement strategies, \hydra{} enables effective scoring across both dynamic and static candidates.
    
    \item \hydra{} demonstrates superior planning performance and generalization to out-of-domain data on the Navhard benchmark~\cite{cao2025pseudo}.
    With model ensembling, our sensor-based \hydra{}—the winning entry of the Navsim v2 Challenge—approaches the performance of the state-of-the-art planner PDM-Closed~\cite{dauner2023parting}, which operates on ground-truth perception and is unaffected by degraded sensor inputs.
\end{enumerate}

\newcommand{\pred}[0]{$\mathcal{S}^m_i$}
\newcommand{\gt}[0]{$\hat{\mathcal{S}}^m_i$}

\section{The Three Pillars of \hydra}

\label{sec:solution}

\subsection{Trajectory Generator}
\label{subsec:dp}

To obtain high-quality dynamic trajectory proposals during inference, 
diffusion models have become a popular practice in autonomous driving for its ability to generate multi-modal trajectory candidates~\cite{jiang2023motiondiffuser, chi2023diffusion, yang2024diffusion, westny2024diffusion, liao2024diffusiondrive, li2025finetuning, xing2025goalflow, zheng2025diffusion}.
We adopt a Diffusion Policy-based trajectory generator following~\cite{chi2023diffusion} to produce multiple trajectory proposals.
This sub-network is composed of an image backbone to extract image features, a BEV encoder where BEV queries attend to image features through a Transformer Encoder~\cite{vaswani2017attention}, and a Diffusion Transformer that generates $N$ trajectory proposals $\mathcal{V}_{dp}$ conditioned on the BEV features.

During training, we follow Transfuser~\cite{chitta2022transfuser} and include a BEV segmentation head to provide supervision on the BEV features.
For the Diffusion Transformer, we apply first-order differentiation to the ground-truth trajectory waypoints to normalize the input and use the
DDPM scheduling strategy~\cite{ho2020denoising} for denoising ground-truth trajectories.

\subsection{Scorer with Vocabulary Generalization}
\label{subsec:dense}
While diffusion-based trajectory generators provide fine-grained trajectory proposals, they remain limited in their ability to capture the full breadth of possible driving scenarios.
To achieve robust generalization, we introduce a novel vocabulary generalization technique that trains the model to effectively evaluate diverse trajectory distributions, even those not seen during training.

We propose the Generalized Vocabulary Scorer \textbf{\hydra-Dense}, which builds upon Hydra-MDP~\cite{li2024hydra} but introduces critical innovations in trajectory evaluation. The architecture consists of an image backbone, a trajectory tokenizer that encodes candidates into feature representations, and a Transformer Decoder~\cite{vaswani2017attention} that models complex interactions between trajectory and image tokens.

Our key innovation is twofold: First, we deliberately train on a super-dense trajectory vocabulary ($\mathcal{V}_{XL}$ with 16,384 distinct trajectories) that comprehensively covers the trajectory space, while inferencing on a smaller vocabulary ($\mathcal{V}_{L}$ with 8,192 trajectories)—forcing the model to develop generalizable representations. Second, we apply vocabulary dropout to $\mathcal{V}_{XL}$ during training, randomly removing half of the trajectories in each batch. This serves multiple purposes: (1) it aligns the number of trajectory tokens during training and inference, (2) it creates intentional distribution shifts that improve robustness, and (3) it acts as an effective regularizer against overfitting to specific trajectory patterns.

This vocabulary generalization technique enables our model to effectively score both static trajectory vocabularies and dynamically generated proposals without requiring dedicated training on both types—a capability previous approaches have struggled to achieve.

\begin{table*}[htbp]
\small
\centering
\begin{tabular}{l|cc|cc|ccc}

    \toprule
    Method 
    & Img. Resolution
    & Backbone
    & Training Vocab.
    & Inference Vocab.
    & $\text{EPDMS}_{1}$
    & $\text{EPDMS}_{2}$
    & $\text{EPDMS}$  \\
    \midrule
    DP w/o Scorer & $512\times 2048$ & V2-99 & - & $\mathcal{V}_{dp}$ (Random)  & 60.9 & 40.8 & 25.6 \\
    \midrule
    \midrule
    \multirow{5}{*}[-0.0ex]{\hydra-Dense} & $512\times 2048$ & EVA-ViT-L & $\mathcal{V}_{XL}$ & $\mathcal{V}_{dp}$   & 76.6 & 48.6 & 36.7 \\
      & $512\times 2048$& EVA-ViT-L& $\mathcal{V}_{XL}$ & $\mathcal{V}_{XL}$     & \textbf{78.1} & 50.2 & 39.7 \\

      & $512\times 2048$& EVA-ViT-L& $\mathcal{V}_{XL}$ & $\mathcal{V}_{dp}\cup \mathcal{V}_{XL}$     & 77.4 & 52.7 & 40.8 \\
      & $512\times 2048$& EVA-ViT-L&$\mathcal{V}_{XL}$ & $\mathcal{V}_{dp}\cup \mathcal{V}_{L}$   & 71.8 & 57.3 & 42.0 \\
      & $512\times 2048$& EVA-ViT-L&$\mathcal{V}_{XL} (\text{Dropout})$ & $\mathcal{V}_{dp}\cup \mathcal{V}_{L}$   & 73.1 & \textbf{59.0} & \textbf{43.4} \\
    \midrule
    \midrule
    Baseline Scorer~\cite{li2024hydra}  & $256\times 1024$& ViT-L  & $\mathcal{V}_{L}$ & $\mathcal{V}_{dp}\cup \mathcal{V}_{L}$   & 73.4 & 54.1 & 40.6 \\
    \hydra-Aug & $256\times 1024$ & ViT-L & $\mathcal{V}_{L}$ & $\mathcal{V}_{dp}\cup \mathcal{V}_{L}$& 75.0 & 56.9 & \textbf{43.4}   \\
\bottomrule
\end{tabular}
\vspace{-2mm}
\caption{\textbf{Roadmap to Generalized Trajectory Scoring.
}
Random denotes that we randomly select a trajectory from $\mathcal{V}_{dp}$ during inference.
V2-99~\cite{lee2019energy} is pretrained from DD3D~\cite{park2021pseudo}, 
EVA-ViT-L~\cite{fang2023eva} is initialized from StreamPETR~\cite{wang2023exploring}, and 
ViT-L~\cite{dosovitskiy2020image} is from Depth Anything~\cite{yang2024depth}. 
}
\vspace{-0.2in}
\label{table:abl}
\end{table*}

\begin{figure}
    \centering
        \includegraphics[width=\linewidth]{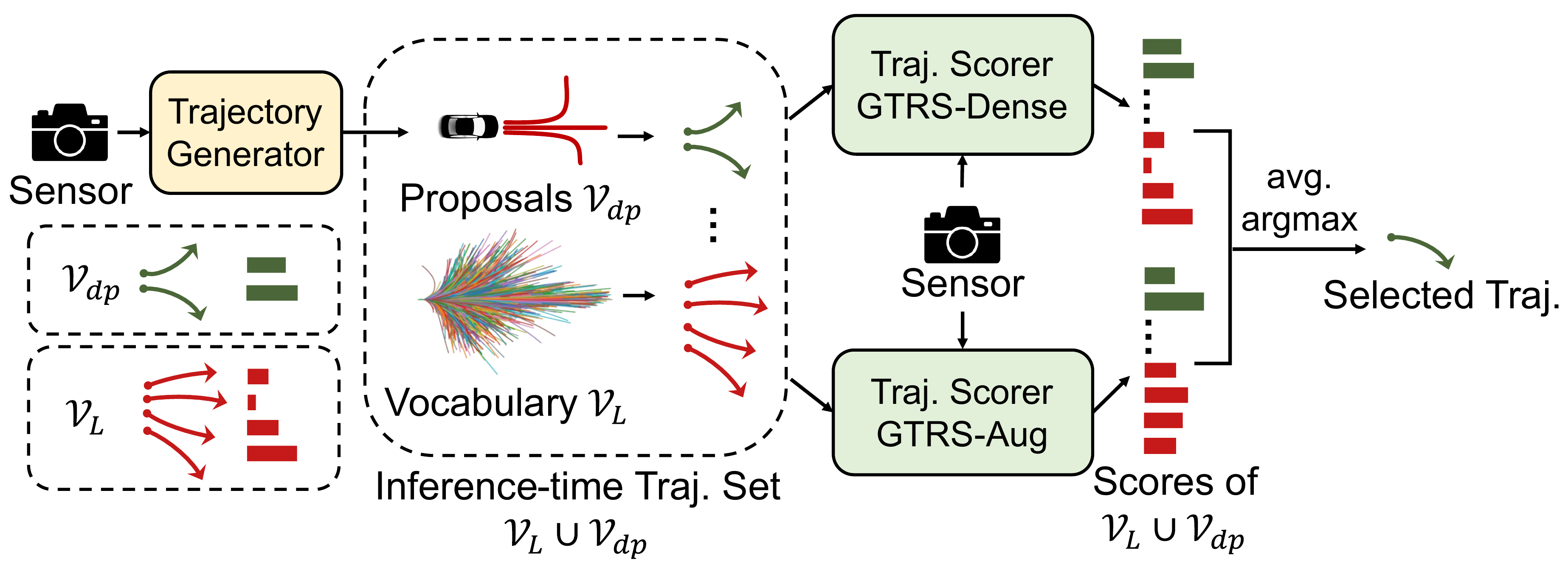}
    \vspace{-0.25in}

    \caption{\textbf{The Inference Pipeline of \hydra.}}
    \vspace{-0.25in}
\label{fig:inference}
\end{figure}

\subsection{Sensor-augmented Scorer with Refinement}
\label{subsec:aug}

To further enhance model robustness across diverse and out-of-domain environments, we develop a systematic sensor augmentation strategy, along with a refinement training mechanism. 
This approach focuses on two critical challenges: handling perceptual distribution shifts in sensor data and distinguishing between subtly different trajectory options in safety-critical scenarios.

First, we introduce structured sensor perturbations by applying controlled 2D horizontal view rotations to the input images. Rather than random augmentation, these perturbations specifically target the model's ability to maintain consistent trajectory evaluation under varying viewing conditions. To maintain label consistency, we apply corresponding transformations to the ground-truths used for training.

Second, we develop a refinement training mechanism focused on fine-grained trajectory discrimination. 
As a training-only module, it incorporates an additional Transformer Decoder that progressively refines trajectory scores for the top-k most promising candidates, enabling the model to capture subtle differences between similar trajectories. The refinement process is guided by a self-distillation framework where an exponential moving average (EMA) copy of the model provides soft supervision signals:
\vspace{-0.03in}
\begin{equation}
        \tilde{y}_{i}^{m} = \hat{y}_{i}^{m} + \mathrm{clip}\left( s_{i, \mathrm{teacher}}^{m} - y_{i}^{m}, -\delta_m, \delta_m \right),
\end{equation}
where $\tilde{y}_{i}^{m}$ represents the refined target score, $\hat{y}_{i}^{m}$ is the ground-truth score, and $s_{i, \mathrm{teacher}}^{m}$ is the teacher model's prediction. The clipping parameter $\delta_m$ ensures the refined targets remain within a reasonable range of the ground truth.

Together, these strategies enable \textbf{\hydra-Aug} to perform robustly in challenging out-of-domain settings without domain-specific adaptation.
\vspace{-0.1in}
\section{Inference-time Integration}

After training the sub-networks described above, we combine the trajectory generator and one of the trajectory scorers (\ie \hydra-Dense, \hydra-Aug) during inference, as illustrated in Fig.~\ref{fig:inference}.
The dynamic proposals generated by the generator $\mathcal{V}_{dp}$ are appended to the inference vocabulary $\mathcal{V}_L$, and the combined set $\mathcal{V}_{dp}\cup\mathcal{V}_{L}$ is tokenized and scored by the scorer.

This sequential integration of dynamic proposals at inference time, rather than during training, is a deliberate design choice that leverages the strengths of both approaches. 
By training solely on a diverse static vocabulary, the scorer develops robust generalization abilities across a wide range of trajectory patterns. 
Then, at inference time, the diffusion-based generator provides fine-grained, context-aware trajectories specifically tailored to the current scene. 
Meanwhile, it avoids the computational overhead and potential instability of integrating diffusion sampling into the training loop, while still benefiting from the precision of dynamically generated trajectories during deployment.

\begin{table*}[htbp]

\scriptsize
\centering
\begin{tabular}{l| c|c |c| c c c c c c c c c |c}

    \toprule
    Method 
    & Img. Resolution
    & Backbone
    & Stage
    & $\text{NC}$
    & $\text{DAC}$
    & $\text{DDC}$
    & $\text{TLC}$
    & $\text{EP}$
    & $\text{TTC}$
    & $\text{LK}$ 
    & $\text{HC}$ 
    & $\text{EC}$ 
    & $\text{EPDMS}$  \\
    \midrule
    
    PDM-Closed~\cite{dauner2023parting}  & GT Perception & - &   \makecell{Stage 1 \\ Stage 2} & \makecell{94.4 \\ 88.1} & \makecell{98.8 \\ 90.6} & \makecell{100 \\ 96.3} & \makecell{99.5 \\ 98.5} & \makecell{100 \\ 100} & \makecell{93.5 \\ 83.1} & \makecell{99.3 \\ 73.7} & \makecell{87.7 \\ 91.5} & \makecell{36.0 \\ 25.4} & 51.3    \\
    \midrule
    \midrule
    
    LTF~\cite{chitta2022transfuser} & $256\times 1024$ & ResNet34 &   \makecell{Stage 1 \\ Stage 2} &
    \makecell{96.2 \\ 77.7} & \makecell{79.5 \\ 70.2} & \makecell{99.1 \\ 84.2} & \makecell{99.5 \\ 98.0} & \makecell{\textbf{84.1} \\ \textbf{85.1}} & \makecell{95.1 \\ 75.6} & \makecell{94.2 \\ 45.4} & \makecell{97.5 \\ 95.7} & \makecell{\textbf{79.1} \\ \textbf{75.9}} & 23.1    \\
    \midrule
    
    \multirow{6}{*}{\hydra-Dense} & \multirow{6}{*}{$512\times 2048$} & V2-99&   \makecell{Stage 1 \\ Stage 2}  &
    \makecell{98.7 \\ 91.4} & \makecell{95.8 \\ 89.2} & \makecell{99.4 \\ 94.4} & \makecell{99.3 \\ 98.8} & \makecell{72.8 \\ 69.5} & \makecell{98.7 \\ 90.1} & \makecell{95.1 \\ 54.6} & \makecell{96.9 \\ 94.1} & \makecell{40.4 \\ 49.7} & 41.7    \\
    \cmidrule{3-14}

     &  & EVA-ViT-L&   \makecell{Stage 1 \\ Stage 2}  &
    \makecell{97.6 \\ 91.9} & \makecell{95.8 \\ 91.3} & \makecell{99.7 \\ 92.7} & \makecell{\textbf{99.8} \\ 99.0} & \makecell{77.2 \\ 72.7} & \makecell{97.8 \\ 90.4} & \makecell{95.3 \\ 53.8} & \makecell{97.3 \\ 94.1} & \makecell{46.7 \\ 41.6} & 43.4    \\
    \cmidrule{3-14}
    
     &  & ViT-L&   \makecell{Stage 1 \\ Stage 2}  &
    \makecell{\textbf{98.9} \\ 91.5} & \makecell{98.2 \\ 90.8} & \makecell{\textbf{99.8} \\ \textbf{94.7}} & \makecell{99.6 \\ 98.5} & \makecell{73.9 \\ 70.8} & \makecell{98.9 \\ 90.1} & \makecell{95.3 \\ 55.4} & \makecell{97.3 \\ \textbf{97.2}} & \makecell{40.0 \\ 54.2} & 45.3    \\
    \midrule

    \multirow{6}{*}{\hydra-Aug} & \multirow{4}{*}{$512\times 2048$} & V2-99&   \makecell{Stage 1 \\ Stage 2}  &
    \makecell{\textbf{98.9} \\ 87.9} & \makecell{95.1 \\ 88.8} & \makecell{99.2 \\ 89.6} & \makecell{99.6 \\ 98.8} & \makecell{76.1 \\ 80.3} & \makecell{99.1 \\ 86.0} & \makecell{94.7 \\ 53.5} & \makecell{\textbf{97.6} \\ 97.1} & \makecell{54.2 \\ 56.1} & 42.1    \\
    \cmidrule{3-14}

     &  & EVA-ViT-L&   \makecell{Stage 1 \\ Stage 2}  &
    \makecell{98.7 \\ 89.5} & \makecell{98.0 \\ 89.6} & \makecell{99.1 \\ 92.9} & \makecell{\textbf{99.8} \\ 98.5} & \makecell{75.9 \\ 78.9} & \makecell{98.7 \\ 86.4} & \makecell{94.7 \\ 55.3} & \makecell{\textbf{97.6} \\ 96.5} & \makecell{49.8 \\ 52.7} & 44.7    \\
    \cmidrule{2-14}
    
     & $256\times 1024$ & ViT-L&   \makecell{Stage 1 \\ Stage 2}  &
    \makecell{97.8 \\ 90.3} & \makecell{97.3 \\ 88.9} & \makecell{98.9 \\ 90.8} & \makecell{99.3 \\ 98.9} & \makecell{77.1 \\ 81.1} & \makecell{98.2 \\ 87.4} & \makecell{95.8 \\ 54.2} & \makecell{\textbf{97.6} \\ 95.1} & \makecell{50.2 \\ 48.3} & 43.4    \\
    \midrule
    
    GTRS-E-Lite & $512\times 2048$ & EVA-ViT-L &   \makecell{Stage 1 \\ Stage 2}  &
    \makecell{98.2 \\ 90.6} & \makecell{98.9 \\ 91.7} & 
    \makecell{99.6 \\ 93.6} & \makecell{99.6 \\ 98.5} & 
    \makecell{75.8 \\ 77.1} & \makecell{\textbf{98.4} \\ 89.0} & \makecell{\textbf{96.9} \\ \textbf{56.1}} & \makecell{97.3 \\ 96.3} & \makecell{53.3 \\ 47.8} & 46.6\\  
    \midrule
    
    GTRS-E & * & *&   \makecell{Stage 1 \\ Stage 2}  &
    \makecell{\textbf{98.9} \\ \textbf{92.3}} & \makecell{\textbf{99.3} \\ \textbf{93.3}} & \makecell{\textbf{99.8} \\ 94.6} & \makecell{\textbf{99.8} \\ \textbf{99.2}} & \makecell{75.2 \\ 73.1} & \makecell{\textbf{98.4} \\ \textbf{91.2}} & \makecell{96.0 \\ 53.9} & \makecell{\textbf{97.6} \\ 96.7} & \makecell{51.6 \\ 56.8} & \textbf{49.4}\\
\bottomrule
\end{tabular}
\vspace{-0.1in}
\caption{\textbf{Performance on the Navhard Benchmark.}
Backbone settings follow Tab.~\ref{table:abl}.
GTRS-E-Lite ensembles GTRS-Dense and GTRS-Aug with EVA-ViT-L.
The challenge-winning entry \textbf{GTRS-E} ensembles all six models from GTRS-Dense and GTRS-Aug.}
\label{table:result}
\vspace{-0.2in}

\end{table*}

\vspace{-0.1in}
\section{Experiments}

\subsection{Dataset and metrics}
The Navsim dataset~\cite{Dauner2024NEURIPS} is designed for evaluating end-to-end driving systems, addressing prior limitations in benchmarking~\cite{li2023ego}. 
The Navsim v2 Challenge introduces a new split, Navhard, which features difficult real-world scenarios alongside their synthetic continuations generated using 3DGS. 
Nevertheless, we observe that the synthetic data exhibits sub-optimal quality, which often contains artifacts such as distortion and blurring and may impair the performance of sensor-based planners.
The Navsim v2 challenge evaluates end-to-end models based on the extended PDM Score (EPDMS)~\cite{li2025hydrapp}, an extension of the PDM Score~\cite{Dauner2024NEURIPS} by aggregating multiple rule-based metrics~\footnote{https://github.com/autonomousvision/navsim/blob/main/docs/metrics.md}.
Finally, it uses a two-stage scoring pipeline
to aggregate the metrics on real-world data (Stage 1) and synthetic data (Stage 2).

\subsection{Implementation Details}
We train all models on the Navtrain split with 24 NVIDIA A100 GPUs, while the Navhard split and other synthetic sensor data are not used for training.
Training is conducted for 20 epochs with a total batch size of 528 by default, while the training lasts 50 epochs for the trajectory generator.
The learning rate and weight decay are $2\times10^{-4}$ and 0.0. 
We concatenate the frontal view with center-cropped front-left and front-right views to form the input image.
For the trajectory generator DP, we formulate a similar input image from the back-view, back-right view, and back-left view for BEV construction.
Finally, we use 100 denoising steps with the DDPM scheduler~\cite{ho2020denoising} and generate 100 proposals in $\mathcal{V}_{dp}$.

\subsection{Roadmap to Generalized Trajectory Scoring}
\label{subsec:roadmap}
\noindent \textbf{Trajectory Vocabulary Generalization.}
We evaluate \hydra-Dense with various inference vocabularies (Tab.~\ref{subsec:roadmap}). Notably, though it is trained only on the super-dense static vocabulary $\mathcal{V}_{XL}$, the scorer generalizes well to unseen dynamic proposals in $\mathcal{V}_{dp}$ (EPDMS: 36.7), demonstrating strong zero-shot generalization by outperforming the generator with random selection substantially (+ 11.1 EPDMS).
When combining $\mathcal{V}_{dp}$ with $\mathcal{V}_{XL}$, performance improves by +1.1 EPDMS over $\mathcal{V}_{XL}$, confirming the complementary benefit of dynamic proposals at inference. 
Interestingly, $\mathcal{V}_{dp}\cup\mathcal{V}_L$ outperforms $\mathcal{V}_{dp}\cup\mathcal{V}_{XL}$, likely because the reduced vocabulary complexity leads to better generalization in out-of-domain synthetic data.
Finally, applying dropout to $\mathcal{V}_{XL}$ during training yields the best performance (EPDMS: 43.4), showing that vocabulary dropout enhances generalization significantly.

\noindent \textbf{Sensor Augmentation with Refinement.}
Finally, we evaluate \hydra-Aug, which incorporates sensor augmentation and refinement training. This model achieves the same top-level performance (EPDMS: 43.4) as the best \hydra-Dense variant, surpassing the baseline scorer~\cite{li2024hydra} by a large margin (+2.8 EPDMS). 
This confirms that augmentations and refinement training greatly improve  trajectory scoring.

\subsection{Main Results}
As shown in Tab.~\ref{table:result}, our \hydra{} variants achieve significant improvements over the LTF baseline~\cite{chitta2022transfuser}. 
By scaling the image backbone to ViT-L~\cite{dosovitskiy2020image} and EVA-ViT-L~\cite{fang2023eva}, our best single model achieves 45.3 EPDMS on the Navhard Benchmark. 
Further, \hydra-E-Lite, an ensemble of \hydra-Dense with \hydra-Aug during scoring, achieves 46.6 EPDMS.
Our challenge-winning entry \hydra-E, an ensemble of all six variants, reaches 49.4 EPDMS, approaching the performance of PDM-Closed~\cite{dauner2023parting}—a privileged planner that relies on ground-truth perception—despite ours using challenging synthetic sensor input. 
This demonstrates the exceptional generalization capabilities of our approach across both trajectory distributions and perceptual domains.

\label{sec:experiment}

{\small
\bibliographystyle{ieeenat_fullname}
\bibliography{11_references}
}


\end{document}